\documentclass[runningheads]{llncs}
\usepackage[T1]{fontenc}
\usepackage[utf8]{inputenc}
\usepackage{graphicx}
\usepackage{amssymb}
\usepackage{amsmath}
\usepackage{xcolor}
\usepackage{subcaption}
\usepackage{hyperref}
\usepackage{multirow}
\usepackage{booktabs}
\begin{document}
%
\title{Self-Supervised Contrastive Learning for Cardiac MR Sequence Classification}

\author{
Yuli Wang\inst{1} \and
Hyewon Jung\inst{2} \and
Dongshen Peng\inst{3} \and
Yuwei Dai\inst{4} \and
Jing Wu\inst{5} \and
Haoyue Guan\inst{1} \and
Yoko Kato\inst{6} \and
Zhicheng Jiao\inst{7} \and
Yu Sun\inst{2} \and
Ihab Kamel\inst{4} \and
Joao Lima\inst{6} \and
Cheng Ting Lin\inst{1} \and
Harrison Bai\inst{1,4}
}

\authorrunning{Wang et al.} 

\institute{
Department of Radiology and Radiological Science, Johns Hopkins University School of Medicine, Baltimore, MD, 21205, USA \and
Department of Electrical and Computer Engineering, Johns Hopkins University, Baltimore, MD, 21218, USA \and
Department of Computer Science, University of North Carolina at Chapel Hill, Chapel Hill, NC 27599 \and
Department of Radiology, University of Colorado Denver Anschutz Medical Campus, Aurora, CO, 80204, USA \and
Department of Radiology, Second Xiangya Hospital, Central South University, Changsha, Hunan, China \and
Department of Cardiology, Johns Hopkins University School of Medicine, Baltimore, MD, 21205, USA \and
Department of Diagnostic Imaging, Brown University Health, Providence, RI, 02903, USA
\\
\email{harrison.bai@cuanschutz.edu}
}

\maketitle              
\begin{abstract}

Vision Transformer (ViT) models, utilizing self-attention mechanisms, have demonstrated robust generalization capabilities across various vision tasks, including image classification. However, these models, typically pretrained on general public datasets, often lack the specialized domain knowledge necessary for medical imaging applications. In this study, we investigate the adaptation of ViT models, specifically for cardiac magnetic resonance (MR) images, using an in-house dataset. We found that pretrained ViT features do not effectively transfer to the cardiac MR domain. To overcome this limitation, we introduce an adaptation strategy that utilizes image-based self-supervised contrastive learning, demonstrating superior performance compared to traditional supervised training approaches. Moreover, our adapted ViT model exhibits strong generalization to external MR datasets such as BraTS and ADNI. Through ablation studies, we further investigate the impact of batch size and dataset scale on performance. Ultimately, our adapted model achieves classification AUC exceeding 0.75 across the four most common cardiac MR sequences.

\keywords{Cardiac MR \and Sequence Classification \and Self-supervise Contrasctive Learning}
\end{abstract}
\section{Introduction}

Vision Transformer (ViT) models \cite{dosovitskiy2020image,wu2025vision}, leveraging self-attention mechanisms originally developed for natural language processing, have significantly advanced the field of computer vision. By dividing images into patches and processing them sequentially, ViTs effectively capture global contextual information and relationships throughout the entire image. This capability has led to state-of-the-art performance in numerous computer vision tasks, including image classification \cite{chen2021crossvit,zhao2026artificial}, object detection \cite{zhou2020automatically,zhou2025dataset}, semantic segmentation \cite{strudel2021segmenter,hsu2025mri}, and image synthesis \cite{kamran2021vtgan}.

ViT models pretrained on extensive public datasets, such as ImageNet \cite{deng2009imagenet}, typically exhibit strong performance on general-domain tasks involving everyday objects, animals, and natural scenes. However, similar to other general-purpose models, ViTs pretrained on natural images frequently encounter performance limitations when directly applied to specialized domains ,such as clinical/ or edical images \cite{wang2025bi,wang2026diffusionqc}, remote sensing \cite{aleissaee2023transformers}, or microscopic cell classification \cite{wang2022label,guo2026multitask}, due to significant domain differences. Therefore, domain-specific adaptation is crucial for fully exploiting their capabilities.

In this study, we explore the adaptation of ViT models specifically for the sequence classification of cardiac magnetic resonance (MR) images. We demonstrate that a ViT pretrained on ImageNet fails to generalize effectively to cardiac MR images. We evaluate two adaptation strategies: supervised fine-tuning (FT) on labeled data and self-supervised contrastive pretraining (SCP) on image data, followed by brief FT. Our results indicate that SCP combined with brief FT significantly outperforms direct supervised learning, suggesting that self-supervised contrastive approaches enable the model to learn more robust and generalizable domain-specific representations.

Additionally, we show that models pretrained with SCP exhibit strong generalization, achieving high performance on external MR datasets. In our first ablation study, we investigate the effect of batch size during contrastive pretraining. We observe that large batch sizes, typical of original ViT training protocols, are not necessary for effective domain adaptation. Contrary to previous beliefs that larger batch sizes consistently improve performance, we find moderate batch sizes yield optimal results, while excessively large batches can negatively impact accuracy. A second ablation study assesses the dataset size required for contrastive pretraining to surpass supervised fine-tuning performance. Remarkably, we find that SCP achieves superior performance compared to FT even with limited data, using as few as 1200 images (approximately 3\%-10\% of our internal dataset).

Furthermore, to our knowledge, there is currently no literature addressing self-supervised methods for cardiac MR image classification. This paper aims to bridge this gap by introducing an automated framework for classifying multi-parametric cardiac MR images with different sequence types, as detailed in Fig. \ref{fig_pipline}. Our proposed algorithm discriminates among various MR image types, including T1-weighted (T1), T2-weighted (T2), Cine, and Late Gadolinium Enhancement (LGE). We validate the effectiveness of our model using the self-supervised contrastive learning architecture, SimSiam \cite{chen2021exploring}.

\section{Background and Related Work}
\begin{figure}
\includegraphics[width=1\textwidth]{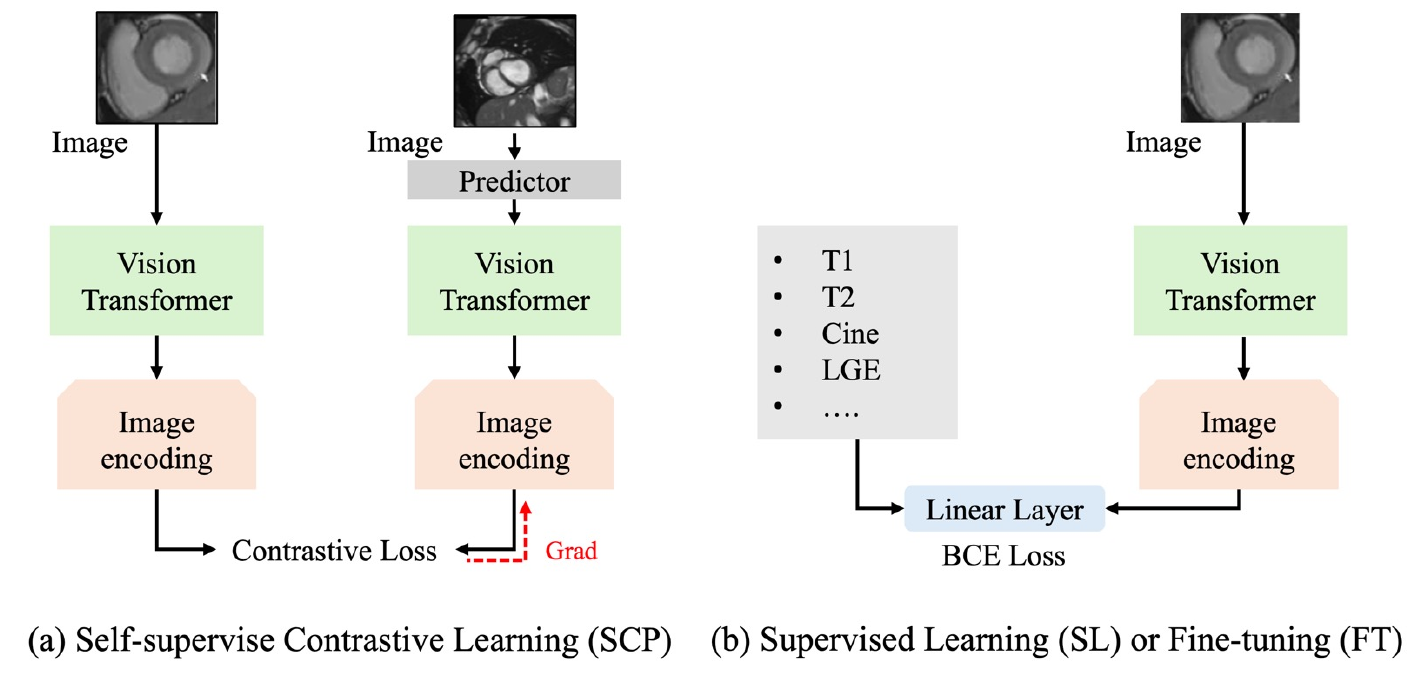}
\centering
\caption{The pipeline of our framework includes (a) self-supervised contractive learning Stage using contrastive loss and SimSiam architecture; (b) supervised learning or supervised fine-tuning stage using BCE loss \cite{su2021dv}.} 
\label{fig_pipline}
\end{figure}

This study explores two training frameworks involving Self-supervised Contrastive Pretraining (SCP) followed by Supervised Fine-tuning (FT) and sole supervised learning (SL). The input comprises a set of images $S$ and their corresponding labels $Y$. A ViT~\cite{beyer2022better} serves as the backbone throughout both stages. During SCP, the network processes pairs of images, generating representations, such as $e_{\text{image1}}$ and $e_{\text{image2}}$. These representations are optimized using a contrastive loss to encourage the model to learn meaningful features without relying on labels. FT then leverages these learned representations to train a classifier using labeled data. Specifically, the encoded image representation $e_{\text{image}}$ is passed through a linear layer, and optimization is performed using the binary cross-entropy (BCE) loss. Fig.~\ref{fig_pipline} outlines the overall training pipeline.

In the FT or SL stage, with the ground truth $y_{ni}$, the model predicts a label $\hat{y}_{ni}$ for each input $s_i \in S$, where $n \in Y$, using a function $f(s_i) = \hat{y}_{ni} $. The loss for each label is computed using binary cross-entropy:

\begin{equation}
L_{\text{BCE}_n}(y_{ni}, \hat{y}_{ni}) = -\left(y_{ni} \log(\hat{y}_{ni}) + (1 - y_{ni}) \log(1 - \hat{y}_{ni})\right)
\end{equation}

During the SCP stage, a batch of $K$ image pairs is sampled, and each image is encoded via the ViT encoder. The loss function encourages alignment between positive pairs, i.e., pairs from the same category, while separating them from negative samples drawn from different categories within the batch. For an anchor sample $x_i$, the matched positive is denoted $x_i^+$, while all mismatched samples are collected as $X^-$. The contrastive loss is computed using the NT-Xent (e.g., the normalized temperature-scaled cross entropy loss)~\cite{oord2018representation,wu2018unsupervised,wang2025dataset} with cosine similarity as the similarity metric:

\begin{equation}
L_{\text{NT-Xent}}(x_i, x_i^+, X^-) = 
- \log \left( \frac{\exp(\text{sim}(x_i, x_i^+))}{\sum_{j=0}^{2K} \exp(\text{sim}(x_i, x_j^-))} \right)
\end{equation}

The total NT-Xent loss is the average of all individual losses of the samples from the batch. 

\section{Method}
All models are evaluated using the mean area under the receiver operating characteristic curve (ROC-AUC), computed across all label categories within each dataset.

\begin{figure}
\includegraphics[width=0.8\textwidth]{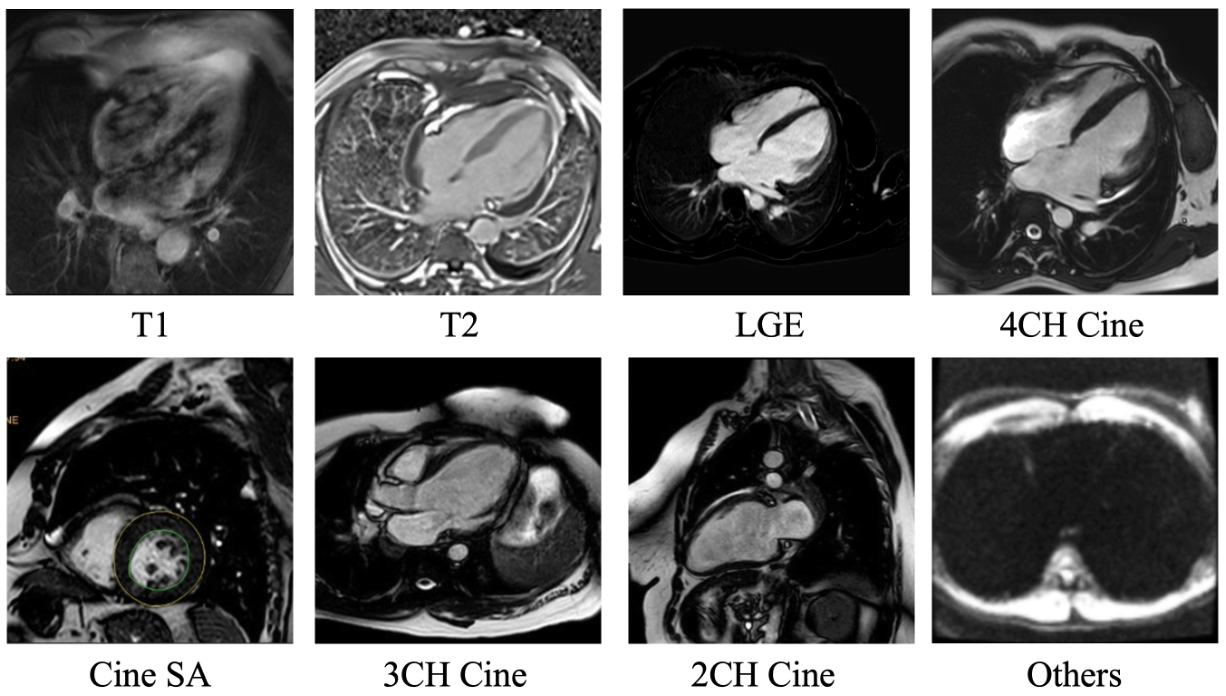}
\centering
\caption{Example of cardiac MR images acquired from different MR sequences. From left to right and top to bottom: T1, T2, LGE, and Cine (including 4-chamber, short-axis, 3-chamber, 2-chamber views).} 
\label{fig_latent_space}

\end{figure}

\subsection{Dataset Curation}\label{dataset}

The in-house internal dataset comprises 4,828 cardiac MR studies, covering four major sequences. The most prevalent sequences include \textbf{T1}, \textbf{T2}, \textbf{Cine}, and \textbf{LGE}. All remaining modalities, such as T1 mapping, T2 mapping, Scout, DWI, and Perfusion, are grouped under the category \textbf{Others}. To construct the working dataset, a random selection of images is made from each study within a MR sequence type, retaining only the central 20\% of sagittal-plane slices. Sequences exhibiting empty or corrupted data, or those with repeated anatomical regions, are removed. From this curated dataset, data from 4,212 patients are split at the patient level into training (70\%, 28,890 images), validation (10\%, 4,110 images), and test (20\%, 8,092 images) sets, ensuring all images from the same study remained within the same subset. For consistency and computational efficiency, all images are resampled to a resolution of 84 × 84 pixels.

To further evaluate the model’s generalizability across various anatomical regions, two publicly available MR datasets are included: the Brain Tumor Segmentation (BraTS) dataset~\cite{menze2014multimodal} and the Alzheimer’s Disease Neuroimaging Initiative (ADNI) dataset~\cite{hinrichs2009spatially}. These datasets were resampled to 80 × 80 pixels to match model input requirements. Due to differences in sequence availability, a separate model is trained for each external dataset to ensure fair performance comparison.

\begin{figure}
\includegraphics[width=0.9\textwidth]{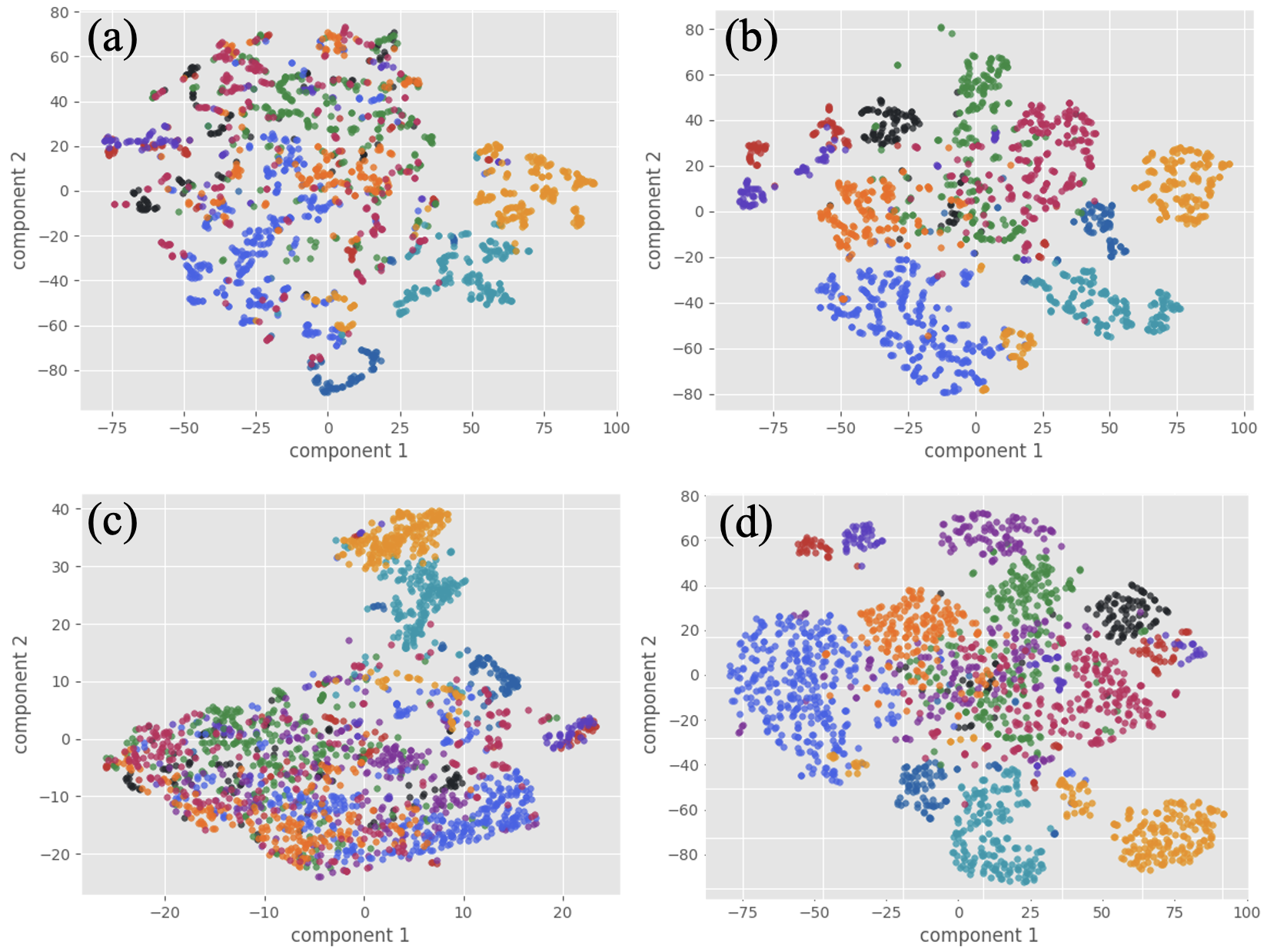}
\centering
\caption{Examples of latent space visualizations from SimSiam during pretraining: (a)–(b) at epoch 5, and (c)–(d) at epoch 150. Visualizations (a) and (c) use a perplexity value of T = 5, while (b) and (d) use T = 100.} 
\label{fig_latent_space}
\end{figure}

\subsection{Pre-training: Self-supervised Contrastive Learning}

Self-supervised contrastive learning operates under the assumption that semantic content remains unchanged under certain image transformations. As a result, different augmented views of the same image, such as horizontal flips, rotations, or elastic deformations in this study, are treated as positive pairs, while all other images in the batch serve as negative pairs. The objective is to bring positive pairs closer in the latent space while pushing negative pairs apart, using a contrastive loss function. SimCLR~\cite{chen2020simple} shows the effectiveness of this approach, achieving performance that surpasses supervised models on ImageNet while using 100 times fewer labels. Despite its success, SimCLR requires extremely large batch sizes to form enough negative pairs, which makes it computationally intensive. To address this issue, SimSiam~\cite{chen2021exploring} introduces a simplified framework based on siamese networks that eliminates the need for negative pairs. This design significantly reduces the demand for a large batch size, while still enabling the model to learn rich, discriminative features. In our study, we adopt SimSiam exclusively for the SCP stage.

\subsection{Self-supervision Contrastive Learning or Supervised Learning}\label{Fine_tuning}

Both SL and SCP are performed using the in-house training dataset described in Section~\ref{dataset}. SCP is further followed by supervised fine-tuning using varying portions of the same dataset, ranging from 1\% to 100\%. To further evaluate model generalizability, two external datasets are curated, BraTS and ADNI, each contributing 220 patient images per available MR sequence (i.e., T1 and T2 sequences). For sequences not present in these external datasets, the original in-house data is used for both fine-tuning and evaluation.

\subsection{Ablation Studies}

The \textit{first ablation study} investigates the effect of batch size on the performance of SCP by varying the number of negative samples. Batch sizes ranging from 4 to 1,024 are tested. Using a ViT backbone initialized with ImageNet pretraining, models are trained on the in-house cardiac MR dataset under the SimSiam framework. 
The \textit{second ablation study} explores how SCP performs with limited data. Dataset sizes are progressively reduced from 100\% to as low as 1\%. All pretraining is conducted with 200 epochs, followed by fine-tuning for either 10 or 50 epochs. 
In the \textit{third ablation study}, we evaluate the model’s generalizability in a zero-shot setting using two external datasets, BraTS and ADNI. 

\section{Results}

\subsection{Self-supervision Contrastive Learning Compared to Supervised Learning}

Table~\ref{results_table} (a) presents a comparison between supervised learning (SL) and self-supervised contrastive pretraining (SCP). When using pretrained ViT models from prior work~\cite{dosovitskiy2020image,dehghani2023patch} with a linear probe, performance is suboptimal, with AUC scores around 35. This suggests that features learned from general-purpose ViTs do not transfer effectively to the cardiac MR domain. Applying SL directly on the in-house dataset significantly improves the AUC to approximately 74. Notably, SCP further enhances performance, achieving AUC scores of 76 and 78 with just 10 and 50 epochs of fine-tuning, respectively. These findings underscore the value of SCP in capturing more transferable and domain-relevant features, improving model generalizability even with minimal label supervision.

\begin{table}[h]
\centering
\caption{Comparison of self-supervised contrastive pretraining (SCP), supervised learning (SL), and supervised fine-tuning (FT). AUC represents the ROC-AUC multiplied by 100.}
\vspace{0.5em}

\begin{subtable}[t]{0.48\textwidth}
\centering
\caption{In-house Cardiac MR dataset}
\begin{tabular}{@{}lllllll@{}}
\toprule
\textbf{Model} & \textbf{Type} & \textbf{AUC} \\
\midrule
\cite{dosovitskiy2020image} & Pre-trained & ~35.6 \\
\cite{dehghani2023patch} & Pre-trained & ~39.4 \\
Ours & SCP + FT 10 Eps & ~76.5 \\
Ours & SCP + FT 50 Eps & ~78.4 \\
Ours & SL 10 Eps & ~75.8 \\
Ours & SL 50 Eps & ~74.3 \\
\bottomrule
\end{tabular}
\end{subtable}
\hfill
\begin{subtable}[t]{0.48\textwidth}
\centering
\caption{External datasets, BraTS and ADNI}
\begin{tabular}{@{}llll@{}}
\toprule
\textbf{Dataset} & \textbf{Model} & \textbf{Type} & \textbf{AUC} \\
\midrule
BraTS & Ours & SCP + FT 50 Eps & ~68.4 \\
BraTS & Ours & SL 50 Eps & ~50.4 \\
ADNI & Ours & SCP + FT 50 Eps & ~67.3 \\
ADNI & Ours & SL 50 Eps & ~50.5 \\
\bottomrule
\end{tabular}

\end{subtable}
\label{results_table}
\end{table}

The evaluation on external datasets, BraTS and ADNI, demonstrates the generalizability of the SCP approach. As shown in Table~\ref{results_table} (b), SCP followed by 50 epochs of fine-tuning achieves AUC scores of 68.4 on BraTS and 67.3 on ADNI. In contrast, the fully SL approach yields significantly lower AUCs of 50.4 and 50.5, respectively. These consistent improvements across two independent datasets suggest that the representations learned through SCP transfer effectively beyond the original cardiac MR domain, underscoring the robustness and adaptability of the model across diverse anatomical regions.

\subsection{Batch Size Ablation}

\begin{figure}
\includegraphics[width=1\textwidth]{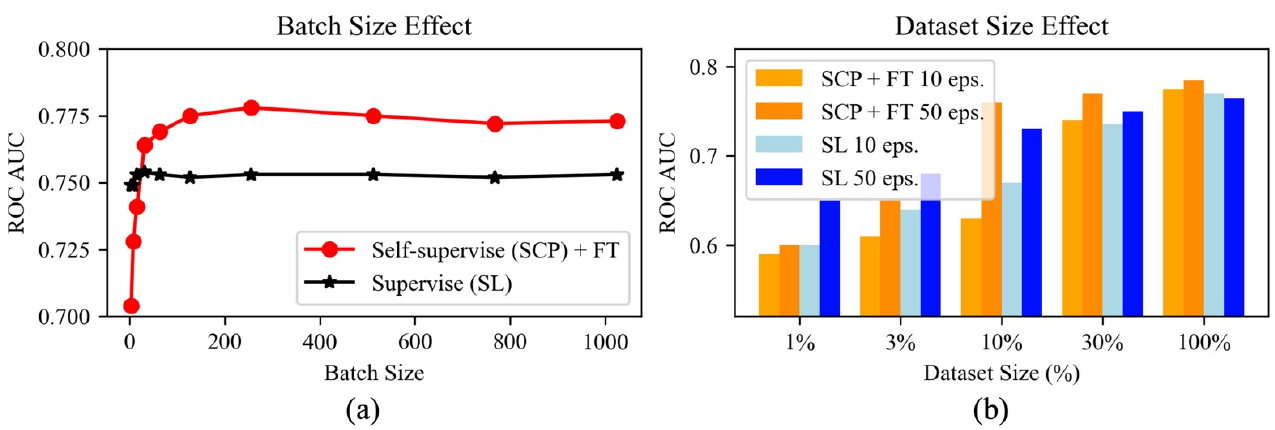}
\centering
\caption{Results for batch size ablation studies and dataset size effects. (a) Batch size ablation studies for self-supervised contrastive learning and supervised learning. Our results consistently show that a batch size of 256 yields the best performance. (b) Performance remains consistent across varying dataset sizes, with SCP + 50 epochs of fine-tuning consistently outperforming the 10-epoch setting for dataset sizes larger than 3\%.} 
\label{fig_comparison}
\end{figure}

Fig.~\ref{fig_comparison} (a) presents the results of the batch size ablation study. To determine the optimal configuration for cardiac MR sequence classification applications, batch sizes ranging from 4 to 1,028 were evaluated under the self-supervised pretraining followed by fine-tuning (SCP + FT) paradigm. While smaller batch sizes result in a slight performance decline, AUC values consistently remain above 70. The highest performance is observed with a batch size of 256, which proves optimal across multiple fine-tuning durations and dataset proportions. In contrast, further increasing the batch size beyond this point leads to a decline in AUC, suggesting the presence of an upper threshold beyond which contrastive learning becomes less effective. These results highlight the critical role batch size plays in shaping the quality of learned representations during contrastive pretraining.


\subsection{Dataset Size Ablation}

Fig.~\ref{fig_comparison} (b) presents the results of the dataset size ablation study, showing the advantages of SCP across varying amounts of labeled data. Pretraining on the full dataset using SCP, followed by fine-tuning on progressively smaller subsets, consistently achieves superior performance compared to models trained solely with supervision. Across all dataset sizes of 3\% or more, SCP with 50 epochs of fine-tuning outperforms the 10-epoch setting. The performance gap between the two settings becomes especially pronounced when at least 10\% of the dataset is used, while the difference narrows at both the low (1\%–3\%) and high (100\%) ends. This pattern suggests a threshold around 10\%, beyond which the benefits of extended fine-tuning become more substantial.

\subsection{Classification Performance on External Datasets}

As shown in Table~\ref{results_table} (b), both fully SL and SCP, followed by FT on the original dataset, exhibit a decline in classification performance when evaluated on external datasets. Despite this performance drop, the SCP approach consistently outperforms the SL. This improvement is likely attributed to the generalizability provided by self-supervised pretraining, which enables the model to transfer more robust features across different organic domains.

\section{Conclusion and Discussion}

Self-supervised contrastive learning, specifically SimSiam, proves effective for cardiac MR sequence classification, demonstrating strong performance across various anatomical regions and imaging modalities. The SCP framework generally outperforms supervised learning with labels, highlighting the power of image-based self-supervision over label reliance in our cardiac MR classification task. These results underscore the potential of self-supervised approaches to improve model generalizability while reducing dependence on large labeled datasets, which is a critical advantage in medical imaging.

We conduct ablation studies on batch size and dataset size, identifying important factors that influence model performance. However, further work is needed to assess performance on larger and more diverse datasets and real-world clinical settings to enhance robustness and clinical applicability.

\section{Acknowledgment}
This work was supported in part by the Innovative Project Award from the American Heart Association 25IPA1454088.

\newpage
\bibliographystyle{splncs04}
\bibliography{mybibliography}
\end{document}